\documentclass{ecai}
\usepackage{graphicx}
\usepackage{latexsym}
\usepackage{amsmath}
\usepackage{amsfonts}
\usepackage{hyperref}
\usepackage{multirow}

\begin{document}

\begin{frontmatter}

\title{Joint Multiple Intent Detection and Slot Filling with Supervised Contrastive Learning and Self-Distillation}

\author{\fnms{Nguyen} \snm{Anh Tu}}
\author{\fnms{Hoang} \snm{Thi Thu Uyen}}
\author{\fnms{Tu} \snm{Minh Phuong}}
\author{\fnms{Ngo} \snm{Xuan Bach}\thanks{Corresponding Author}} 

\address{Department of Computer Science,\\
Posts and Telecommunications Institute of Technology, Hanoi, Vietnam\\
\small{\{anhtunguyen446,thuuyenptit\}@gmail.com;\{phuongtm,bachnx\}@ptit.edu.vn}
}

\begin{abstract}
Multiple intent detection and slot filling are two fundamental and crucial tasks in spoken language understanding. Motivated by the fact that the two tasks are closely related, joint models that can detect intents and extract slots simultaneously are preferred to individual models that perform each task independently. The accuracy of a joint model depends heavily on the ability of the model to transfer information between the two tasks so that the result of one task can correct the result of the other. In addition, since a joint model has multiple outputs, how to train the model effectively is also challenging. In this paper, we present a method for multiple intent detection and slot filling by addressing these challenges. First, we propose a bidirectional joint model that explicitly employs intent information to recognize slots and slot features to detect intents. Second, we introduce a novel method for training the proposed joint model using supervised contrastive learning and self-distillation. Experimental results on two benchmark datasets MixATIS and MixSNIPS show that our method outperforms state-of-the-art models in both tasks. The results also demonstrate the contributions of both bidirectional design and the training method to the accuracy improvement. Our source code is available at \href{https://github.com/anhtunguyen98/BiSLU}{https://github.com/anhtunguyen98/BiSLU}.
\end{abstract}
\end{frontmatter}

\section{Introduction}
\begin{figure*}[t]
	\begin{center}
		\includegraphics[width = 17cm]{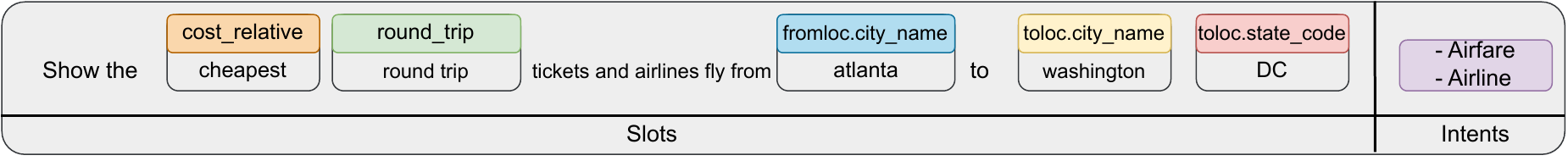}
	\end{center}	
            \vspace{-10pt}
		\caption{An example from the MixATIS dataset.}
	\label{fig:example}
\end{figure*}

Spoken language understanding (SLU) is a core component of task-oriented dialogue systems - an important class of natural language processing (NLP) applications. With the aim of capturing the semantics of user utterances, SLU consists of two main tasks: 1) multiple \textit{intent detection}, which identifies the intents or desires of the user in utterances; and 2) \textit{slot filling}, which extracts slots that provide necessary information to fulfill those intents \cite{gangadharaiah-narayanaswamy:2019}. Figure \ref{fig:example} shows an annotated sample from the MixATIS corpus \cite{qin:2020}, a benchmark dataset widely used in the SLU research community. Given the user utterance ``\textit{Show the cheapest round trip tickets and airlines fly from atlanta to washington DC}'', an SLU component will detect two intents, i.e., airfare and  airline, and extract five slots, i.e., ``cheapest'' (cost\_relative), ``round trip'' (round\_trip), ``atlanta'' (fromloc.city\_name), ``Washington'' (toloc.city\_name), and ``DC'' (toloc.state\_code).  

Traditional approaches to multiple intent detection and slot filling consider them as independent problems, namely semantic classification and sequence labelling, and use a separate model for each task. Those approaches ignore the fact that intents and slots are related to each other. For example, intent airfare often requires some specific types of slots such as fromloc.city\_name and toloc.city\_name. At the same time, slots fromloc.city\_name and toloc.city\_name tend to occur in an utterance with intent airfare or flight. To utilize such relationships between intents and slots, recent approaches rely on joint models that can detect intents and extract slots simultaneously. Thanks to recent advancements in deep learning, various deep neural network-based joint models have been developed and achieved state-of-the-art results in benchmark SLU datasets \cite{cai:2022,chen-coling:2022,ding:2021,song:2022}. 

There are several challenges when designing a successful joint model. First, the accuracy of such a model depends heavily on the way information is transferred between the two tasks. The model should be designed so that the result of one task can be used to correct or improve the result of the other, and vice versa. Existing joint models, however, are often unidirectional. Other models transfer information between two tasks implicitly. Those models cannot fully exploit the relationships between the two tasks to get the improved results. In this work, we propose a bidirectional joint model that explicitly utilizes intent information to extract slots and slot features to detect intents (BiSLU). Given an utterance, our model employs a language model-based encoder to generate its representation and intermediate (soft) intents, which are utilized to extract slots with a biaffine classifier. The model then uses slot features as well as the utterance representation to predict the final intents. 

Another challenge with a joint model is how to train the model effectively. Such a model typically has several outputs, and the information is circulated between the intent and slot components, making the selection of training objectives non-trivial. Traditionally, intent detection and slot filling are cast as supervised learning problems. In this paper, we propose a training framework that includes also contrastive learning and self-distillation. Contrastive learning is a powerful technique for various tasks in different learning scenarios, including self-supervised learning and supervised learning \cite{chen:2020,He:2020,khosla:2020,zhang:2022}. Contrastive learning methods aim to produce better representations of data by maximizing agreement between a data sample (a.k.a. the ``anchor'') and its different augmentations (or views), while maximizing disagreement between the anchor and negative samples, e.g. other samples from the same batch. Following the success of contrastive learning in computer vision, several studies have investigated contrastive learning for NLP tasks \cite{gunel:2021,Rethmeier:2023,ZhangNAACL:2022}. A key advantage of contrastive learning is that machine learning models in different settings can be trained effectively by integrating suitable contrastive losses. In this paper, we propose a method to generate augmentations for the intent detection and slot filling task and integrate the contrastive loss with the original classification loss. 

The final component of our method is self-distillation. Here, we use self-distillation to transfer knowledge from the final intents to the intermediate intents, which leads to an improvement in the joint model’s performance. With contrastive learning and self-distillation, we train our proposed model with a joint loss function consisting of five components: intent loss, slot loss, contrastive intent loss, contrastive slot loss, and self-distillation loss between the intermediate intents and the final intents. 

We verify the effectiveness of the overall method and the contribution of each component in two benchmark datasets: MixATIS and MixSNIPS \cite{qin:2020}. The model achieves a new state-of-the-art in joint multiple intent detection and slot filling with relative error reductions ranging from 3\% to 22\%. The experiments also demonstrate the contribution of contrastive learning and self-distillation components on the accuracy of the final model.

Our main contributions are summarized below:

\begin{enumerate}
    \item We propose a bidirectional joint model for multiple intent detection and slot filling (BiSLU).
    \item We introduce a novel method for training the proposed model effectively using supervised contrastive learning and self-distillation. 
    \item We empirically show the efficacy of BiSLU as well as the proposed training method on two benchmark datasets MixATIS and MixSNIPS. 
\end{enumerate}

In the following, we first review related work on joint multiple intent detection and slot filling in Section 2. We next introduce our proposed method in Section 3, including the bidirectional joint model and the training method. We then describe experimental results and analyses in Section 4. Finally, we conclude the paper in Section 5.  

\section{Related Work}
The interdependence between intent detection and slot filling has been a subject of considerable academic interest, and various models that address both tasks simultaneously within a single framework have been proposed \cite{Haihong:2019,Qin:2019,qin:2021co}. These models, however, exhibit a significant limitation in their intent detection modules, as they are restricted to single-intent utterances, which may be insufficient for real-world applications where multi-intent utterances are prevalent.

Several studies have been conducted for detecting multiple intents of utterances. Gangadharaiah and Narayanaswamy \cite{gangadharaiah-narayanaswamy:2019} pioneered a model that concurrently addresses multiple intent identification and slot filling tasks using an attention-based network. Qin et al. \cite{qin:2020} proposed an adaptive graph interactive framework (AGIF) that leverages a fine-grained approach to integrate multi-intent information into slot filling. Chen et al. \cite{chen:2022} presented a novel self-distillation joint model (SDJN) for multi-intent detection and slot filling. This approach involves mutual intent and slot data sharing for cyclical optimization and employs self-distillation by considering the decoder slots as soft labels for the initial decoder slots.

In recent years, contrastive learning has gained attraction, particularly for self-supervised representation learning, resulting in state-of-the-art performance in unsupervised training of deep image models \cite{chen:2020}. Khosla et al. \cite{khosla:2020} extended contrastive learning to supervised setting, which allows us to leverage label information. In supervised contrastive learning, samples with the same label as the anchor in the batch are considered as positive samples, and the rest is regarded as negative ones. To work with multi-label data, Zhang et al. \cite{zhang:2022} introduced a multi-label contrastive learning framework, which has been shown to be effective in various computer vision tasks. Contrastive learning has been also applied successfully in various NLP tasks, including text classification, sentence embedding, and question answering \cite{karpukhin:2020,kim:2021,suresh:2021,you:2021,zhang:2021}. 

Compared to previous studies, we also develop a joint model that deals with the both tasks simultaneously. However, our method has several significant differences: 1) we introduce a novel architecture for bidirectional joint multiple intent detection and slot filling; 2) we employ supervised contrastive learning and self-distillation to train the proposed joint model effectively. To the best of our knowledge, this is the first attempt to incorporate contrastive learning into SLU. Furthermore, we use a new self-distillation technique, which is different from the one described in Chen et al. \cite{chen:2022}. As shown in experiments, our method is superior to state-of-the-art multi-intent SLU methods.    
\begin{figure}[t]
       \vspace{-10pt}
	\begin{center}
		\includegraphics[width = 0.5\textwidth]{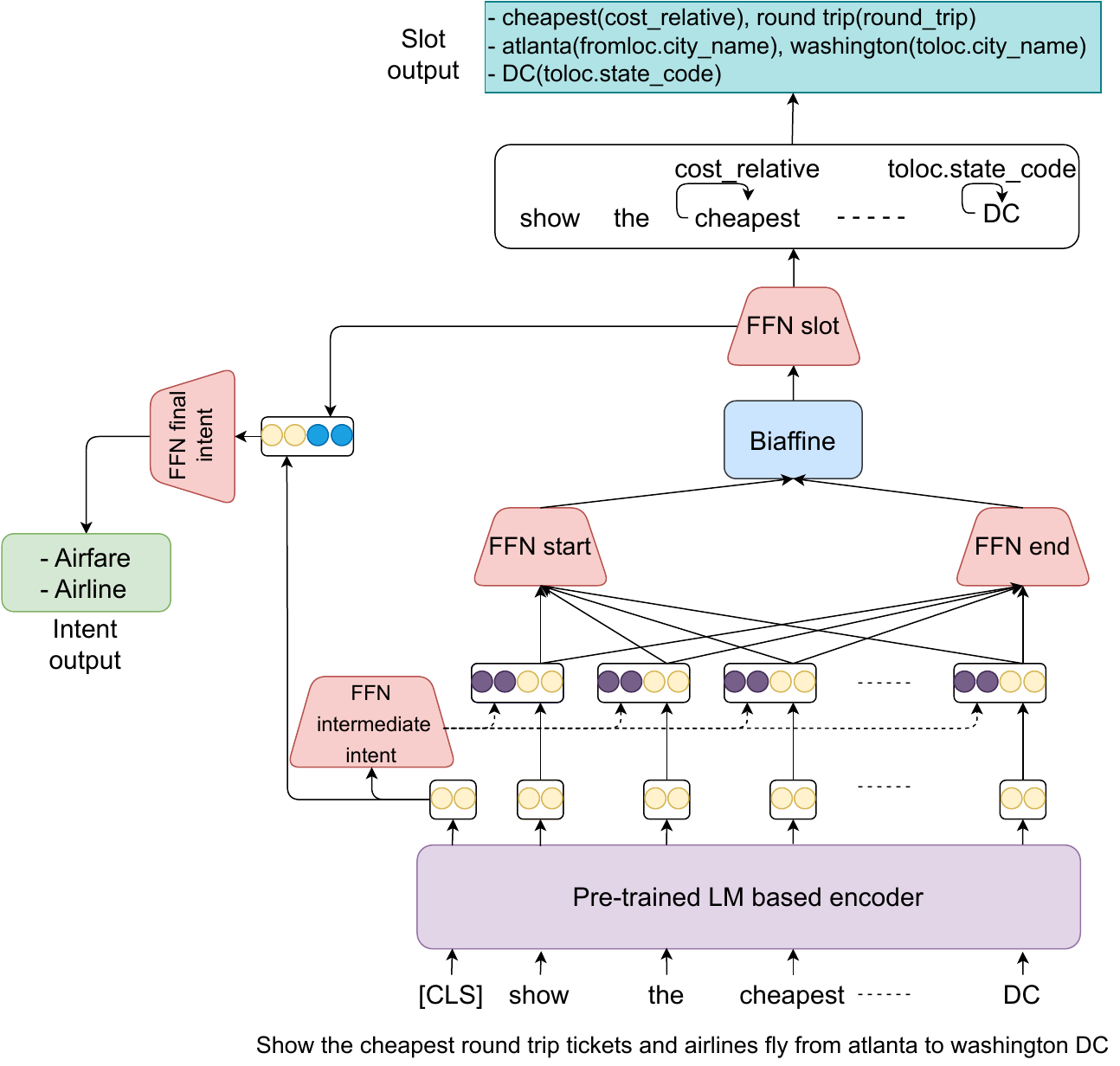}
	\end{center}
	\vspace{-5pt}
		\caption{The architecture of the proposed model.}
	\label{fig:model}
\end{figure}
\section{Our Method}
We first present our bidirectional joint model for SLU (BiSLU) (Section 3.1). We then describe our training method with supervised contrastive learning (Section 3.2), self-distillation (Section 3.3), and a joint training procedure (Section 3.4). 

\subsection{Bidirectional Joint Model}
The architecture of our proposed joint model is illustrated in Figure \ref{fig:model}, consisting of four components: encoder, intermediate intent detection, slot classifier, and final intent detection. Below we describe each component in detail. 

\begin{figure*}[t]
	\begin{center}
		\includegraphics[width = 1\textwidth]{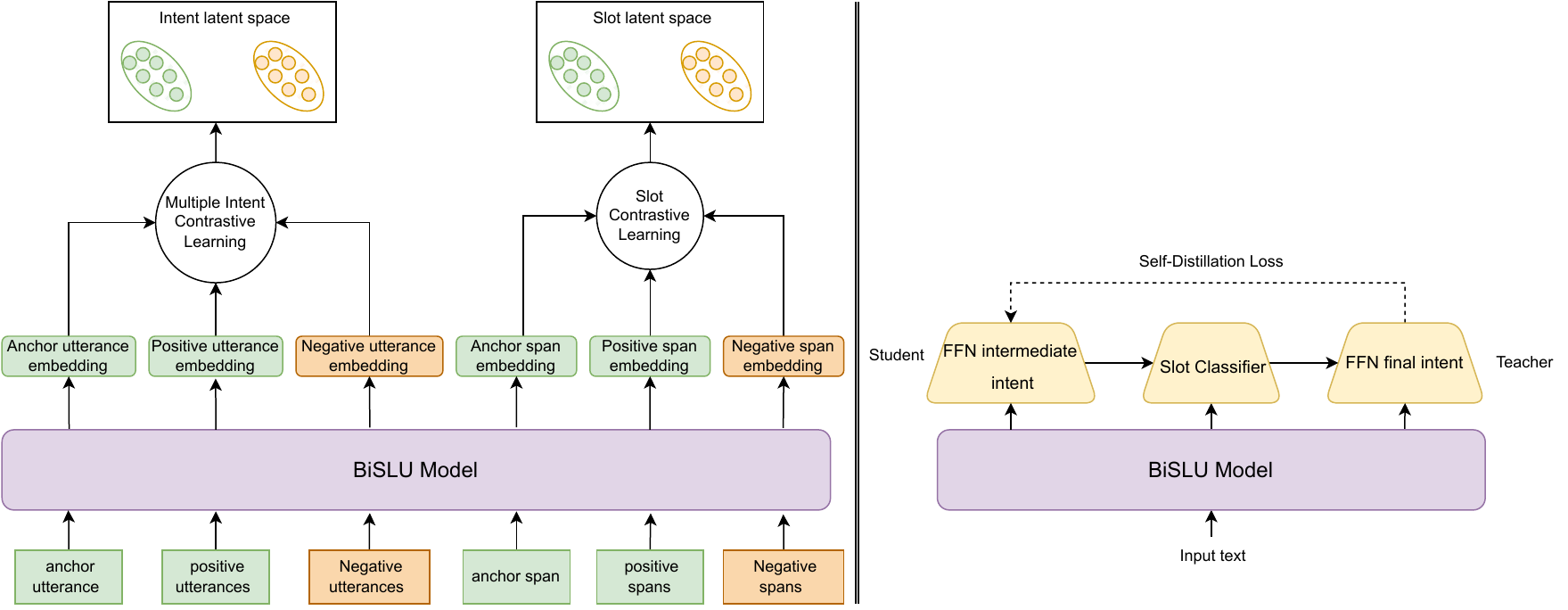}
	\end{center}
	\vspace{-8pt}
		\caption{Supervised contrastive learning (left) and self-distillation (right) for joint multiple intent detection and slot filling.}
	\label{fig:loss}
\end{figure*}

\subsubsection{Encoder}
Given an input utterance consisting of $n$ words $w_1w_2 \ldots w_n$, we prepend a special classification token [CLS], represented by $w_0$, to form the input sequence $w_0w_1w_2 \ldots w_n$. Our encoder utilizes a BERT-base model \cite{Devlin:2019} to generate contextualized word embeddings:
$$\textbf{c}i = BERT\_Encoder(w_{0:n}, i),$$
where $\textbf{c}_i\in \mathbb{R}^d$ $(1 \leq i \leq n)$ corresponds to the embedding of word $w_i$, $\textbf{c}_0$ represents the embedding of the token [CLS] that signifies the entire utterance, and $d$ indicates the embedding size. It is important to note that the encoder segments words into sub-words for efficient rare word modeling. The contextual embedding of a word is obtained by summing up the embeddings of its constituent sub-words.

\subsubsection{Intermediate Intent Detection \& Word Representations}
We feed the embedding $\textbf{c}_0$ into a feed-forward neural network (FFN) to predict intermediate intents: 
$$\textbf{p}= Sigmoid(FFN^{Intermediate\_Intent}(\textbf{c}_0)),$$
where $\textbf{p} \in \mathbb{R}^l$ is a score vector of each intent probability, and $l$ denotes the number of intents. 

To create a word representation, the intermediate intent vector $\textbf{p}$ is concatenated with each contextualized embedding $\textbf{c}_i$ as follows:
$$\textbf{v}_i= \textbf{p}\oplus \textbf{c}_i,$$
where $\textbf{v}_i\in \mathbb{R}^{l+d}$ is the final representation of the $i^{th}$ word, and $\oplus$ denotes the concatenation operation.

\subsubsection{Slot Classifier}
To extract slots, we employ a biaffine classifier, which provides a global view of the input sequence and, therefore, is effective for sequence tagging and related tasks \cite{Li:2019}. We use two feed-forward networks, $FFN^{start}$ and $FFN^{end}$, to create different representations for the start/end position of slots. The outputs of $FFN^{start}$ and $FFN^{end}$ at position $i$ are denoted by $\textbf{g}_i^{start}$  and $\textbf{g}_i^{end}$, respectively:
$$\textbf{g}_i^{start} =FFN^{start}(\textbf{v}_i),$$
$$\textbf{g}_i^{end} =FFN^{end}(\textbf{v}_i).$$

For each start-end candidate slot $(i,j)$ $1 \leq i \leq j \leq n$, we apply the biaffine classifier:
\begin{equation}
\begin{aligned}
\textbf{z}_{i,j} & = \text{Biaffine}(\textbf{g}_i^{start},\textbf{g}_j^{end}) \\
                 & = (\textbf{g}_i^{start})^{\top}  \textbf{U} \textbf{g}_j^{end} + \textbf{W}(\textbf{g}_i^{start}\oplus \textbf{g}_j^{end})+\textbf{b}, \nonumber
\end{aligned}
\end{equation}
where $\textbf{U}$, $\textbf{W}$, and $\textbf{b}$ are a $k \times s \times k$ tensor, a $s \times 2k$ matrix, and a bias vector, respectively; $k$ denotes the dimension of the output layers of two FFNs; and $s$ is the dimension of vector $\textbf{z}_{i,j}$, which represents the segment feature going through a slot classifier $FFN^{slot}$:

\begin{equation}
\nonumber
    \textbf{r}_{i,j} = FFN^{slot}(\textbf{z}_{i,j})
\end{equation}
where $\textbf{r}_{i,j} \in \mathbb{R}^{c+1}$, and $c$ denotes the number of slot labels ($c + 1$ because we add a special label for non-slot
segments). Finally, vector $\textbf{r}_{i,j}$ is fed into a softmax function to produce the probability scores:
$$\textbf{q}_{i,j}(t)= \frac{\exp(\textbf{r}_{i,j}(t))}{\sum_{t'=1}^{c+1}{\exp(\textbf{r}_{i,j}(t'))}}.$$
The slot label of segment $(i,j)$ can be determined as:
$$\arg \max_t \textbf{q}_{i,j}(t).$$
Among overlapping predicted slots, if any, we keep only the slot with the highest score and discard the rest.

\subsubsection{Final Intent Detection}
We create a soft slot vector from the output matrix of the slot classifier: 
$$\textbf{h} = Softmax(\sum_{i=1}^n \sum_{j=i}^n \textbf{r}_{i,j})$$
The output vector $\textbf{h} \in \mathbb{R}^{c+1}$ is concatenated with the representation vector of the token [CLS] ($\textbf{c}_0$) to form the input $\textbf{x} \in \mathbb{R}^{d+c+1}$ to predict the final intent: 
$$\textbf{x} = \textbf{c}_0 \oplus \textbf{h},$$
$$\textbf{p}_{final\_intent}=Sigmoid(FFN^{Final\_Intent}(\textbf{x})),$$
where $FFN^{Final\_Intent}$ is a feed forward neural network, and $\textbf{p}_{final\_intent}\in \mathbb{R}^l$ is the score vector of final intents. Recall that $l$ and $c$ are the number of intent labels and the number of slot labels, respectively. To obtain the final intents, we apply a threshold $0 < t_I < 1$ and select all intents $m$ ($1 \leq m \leq l$) whose probability is greater than $t_I$.

\subsection{Intent and Slot Contrastive Learning }

Supervised contrastive learning (SCL) has achieved remarkable success in computer vision and NLP \cite{khosla:2020,zhang:2022,ZhangNAACL:2022,Rethmeier:2023}, aiming to maximize similarities between instances from the same class while minimizing similarities between instances from different classes. In general, supervised contrastive learning consists of two steps:
\begin{enumerate}
    \item \textbf{Positive/negative sample construction}. Given an anchor, i.e., a training sample in a mini-batch, generates/selects positive and negative samples for the anchor.
    \item \textbf{Loss function design}. Builds an appropriate loss function for generated positive/negative samples.  
\end{enumerate}

The idea of our SCL method for joint multiple intent detection and slot filling is shown in Figure \ref{fig:loss}, on the left-hand side. Below we describe those two steps in detail.  

\subsubsection{Positive/Negative Sample Construction}

\begin{itemize}
    \item \textbf{Positive samples:} To create positive samples for a given anchor sample $x = w_1 w_2 \ldots w_n$, we generate positive representations for both the utterance and spans as follows:

    \begin{itemize}
        \item \textbf{Positive utterances}: For a given anchor utterance, we encode the input utterance $V$ times by applying different dropout rates in the encoder. Each of these $V$ encodings will generate a slightly different representation of the input utterance, resulting in a multi-viewed mini-batch. All utterance representations in the multi-viewed mini-batch, (including augmented utterance representations and other utterances which have the same label as the anchor) are considered as positive samples. A positive pair for the utterance will be formed by pairing the anchor utterance representation $\textbf{c}^{cls}$, with a corresponding positive sample denoted by $\textbf{c}^{cls}_p$.

        \item \textbf{Positive spans}: Similarly, we also encode the input utterance $V$ times. For each encoded utterance, we extract span representations based on the output of the biaffine layer and the start/end indices of the spans. The span representation of the anchor span is $\textbf{z}_{i,j}$, where $(i,j)$ are the word indices in the input utterance and $1 \leq i \leq j \leq n$. All span representations within a multi-viewed mini-batch are regarded as positive spans for the anchor span. We form a positive pair for the span by pairing the anchor span representation $\textbf{z}_{i,j}$ with a corresponding positive sample represented by $\textbf{z}_{p(i,j)}$.
        
    \end{itemize} 

    \item \textbf{Negative samples:} For a given anchor utterance, any instance in the multi-viewed mini-batch that has a different intent class from the original sample is selected as a negative sample. Similarly, for a given anchor span, any instance in the multi-viewed mini-batch that has a different span class from the original sample is selected as a negative sample.
\end{itemize}

\subsubsection{Slot Contrastive Loss}
Let $P(i,j)$ and $A(i,j)$ denote the set of all positive samples and the set of all positive and negative samples corresponding to span $(i,j)$, and $I$ is the collection of all the training spans, we define a supervised contrastive loss for slots as follows:

\begin{equation}
\nonumber
    \mathcal{L}_{sf\_scl}=\sum_{(i,j) \in I} \frac{-1}{|P(i,j)|} \sum_{p \in P(i,j)} \log \frac{f\left(\textbf{z}_{i,j}, \textbf{z}_{p(i,j)}\right)}{\sum_{k \in A(i,j)} f\left(\textbf{z}_{i,j}, \textbf{z}_{k(i,j)}\right)}
\end{equation}
where $f\left(\textbf{z}_{i,j}, \textbf{z}_{p(i,j)}\right) = \exp(\textbf{z}_{i,j} \cdot \textbf{z}_{p(i,j)} /\tau)$ calculates the similarity between $\textbf{z}_{i,j}$ and $\textbf{z}_{p(i,j)}$, and $\tau$ denotes the temperature, a scalar to stabilize the calculation. Recall that $\textbf{z}_{i,j}$ and $\textbf{z}_{p(i,j)}$ are vector representations of span $(i,j)$ of the anchor sample and a positive sample, respectively, created from the output matrix of the biaffine layer. Here, $i$ and $j$ are indices, $1 \leq i \leq j \leq n$, and $n$ denotes the length of an input utterance. 

\subsubsection{Multiple Intent Contrastive Loss}
Although supervised contrastive learning can distinguish between multiple positive pairs, it is only designed for single labels. Following \cite{zhang:2022}, we define $M$ as the set of all intent labels, and $m \in M$ is a label in the label set. The loss for a pair of the anchor utterance, indexed by $i$, and a positive utterance of label $m$ can be defined as follows:
$$
L^{\text {pair }}\left(i, p_m^i\right)=\log \frac{f\left(\textbf{c}_i^{cls}, \textbf{c}_{p_m^i}^{cls}\right)}{\sum_{k \in A(i)} f\left(\textbf{c}_{k_m^i}^{cls}, \textbf{c}_i^{cls}\right)}
$$
where $\textbf{c}_i^{cls}$ and $\textbf{c}_{p_m^i}^{cls}$ are the representation vector of the token [CLS] of the anchor sample and positive samples, respectively; and $A(i)$ denotes the set of all the positive and negative samples corresponding to sample $i$. The multi-label contrastive loss for intents can then be defined as follows:

$$\mathcal{L}_{id\_scl}=\sum_{m \in M} \frac{1}{|M|} \sum_{i \in I} \frac{-\lambda_m}{\left|P_m(i)\right|} \sum_{p_m \in P_m} L^{\mathrm{pair}}\left(i, p_m^i\right)$$
where $\lambda_m = F(m)$ is a controlling parameter that applies a fixed penalty to each label; $F$ is a scale function with $m$ (i.e., exp or pow); and $P_m$ denotes the set of all positive samples of the anchor utterance indexed by $i$. 
\subsection{Self-Distillation}

Recall that our joint model conveys information bi-directionally: the intermediate intent probabilities serve as part of the input for the slot filling layer, and the predicted slot labels are then utilized to determine the final intents. Consequently, the quality of the intermediate intent detection module significantly impacts on the slot filling and final intent detection modules. We propose a self-knowledge distillation method within the joint training model, where the teacher model is the final intent detection layer, and the student model is the intermediate intent detection layer. Our idea is shown in Figure \ref{tab:loss}, on the right-hand side. We compute the representative distance by leveraging the hidden states of both the final intent detection and intermediate intent detection layers. During the training process, we aim to minimize this representative distance between the two hidden states ($\textbf{h}_{intermediate\_intent}$ and $\textbf{h}_{final\_intent}$) based on the logits-based distillation approach.

Conventional logits-based distillation approaches typically minimize the Kullback-Leibler (KL) divergence between the predicted probabilities, that is, the logits after applying the softmax function, of the teacher and student models. However, directly applying this method to multi-label learning (MLL) scenarios is challenging due to the inherent assumption that the predicted probabilities of all classes should sum to one, an assumption that is seldom valid in MLL cases. To address this limitation, we draw inspiration from the one-versus-all reduction concept and propose a multi-label distillation loss. This approach decomposes the original multi-label task into several binary classification problems and aims to minimize the divergence between the binary predicted probabilities of the teacher and student models. The formal definition of self-distillation loss is as follows:
$$\textbf{h}_{S} = \textbf{h}_{intermediate\_intent} = FFN^{Intermediate\_Intent}(\textbf{c}_0),$$
$$\textbf{h}_{T} = \textbf{h}_{final\_intent} = FFN^{Final\_Intent}(\textbf{x}),$$
$$\textbf{p}_{S} = Sigmoid(\textbf{h}_{S}),$$
$$\textbf{p}_{T} = Sigmoid(\textbf{h}_{T}),$$
$$\mathcal{L}_{sd} = KL(\textbf{p}_{S}||\textbf{p}_{T}) + KL(1- \textbf{p}_{S}||1 - \textbf{p}_{T})$$
where $KL(P||Q)$ denotes the Kullback-Leibler divergence of $P$ from $Q$, two discrete probability distributions defined on the same sample space $\mathcal{X}$:
$$KL(P||Q)=\sum_{x \in \mathcal{X}}P(x)log(\frac{P(x)}{Q(x)}).$$

\subsection{Joint Training}
We define a joint loss function $\mathcal{L}$ of our bidirectional model as the weighted sum of the intent detection loss $\mathcal{L}_{id}$, the slot filling loss $\mathcal{L}_{sf}$, the slot contrastive loss $\mathcal{L}_{sf\_{scl}}$, the intent contrastive loss $\mathcal{L}_{id\_{scl}}$, and the self-distillation loss $\mathcal{L}_{sd}$:  
$$\mathcal{L}=  \lambda_1 \mathcal{L}_{id} + \lambda_2 \mathcal{L}_{sf} + \lambda_3 \mathcal{L}_{id\_scl} + \lambda_4 \mathcal{L}_{sf\_scl} + \lambda_5 \mathcal{L}_{sd}$$
where $\mathcal{L}_{id}$ and $\mathcal{L}_{sf}$ use the standard cross-entropy loss; and $0 \leq \lambda_i \leq 1$ ($\sum_{i=1}^5{\lambda_i} = 1$) are hyperparameters, which are tuned using a development set in experiments.

\section{Experiments}
\subsection{Data and Evaluation Methods}
We performed experiments using two benchmarks multi-intent SLU datasets, namely MixATIS and MixSNIPS \cite{qin:2020}. They are the multiple intent versions of the ATIS dataset \cite{Hemphill:1990} and the SNIPS dataset \cite{Coucke:2018}, which were created for single intent detection and slot filling. Both multi-intent SLU datasets exhibit a distribution of sentences containing 1-3 intents at proportions of [0.3, 0.5, 0.2]. In the MixATIS dataset, there are 13,162 training utterances, 759 validation utterances, and 828 testing utterances. Meanwhile, the MixSNIPS dataset consists of 39,776 training utterances, 2,198 validation utterances, and 2,199 testing utterances.

We evaluated the performance of different joint models for multiple intent detection and slot filling using three conventional metrics as in \cite{gangadharaiah-narayanaswamy:2019}: intent accuracy, slot F$_1$ score, and sentence-level semantic frame accuracy. 

\begin{table*}[t]
\centering
\caption{Performance comparison on the MixATIS and MixSNIPS datasets. The best values for each column
are shown in bold}
\label{tab:Results}
\begin{tabular}{|l|c|c|c|c|c|c|}
\hline
\multirow{2}{*}{\textbf{Model}} & \multicolumn{3}{|c|}{\textbf{MixATIS}} &	\multicolumn{3}{|c|}{\textbf{MixSNIPS}} \\
\cline{2-7}
& \textbf{Intent} &	\textbf{Slot} &	\textbf{Sent} & \textbf{Intent} &	\textbf{Slot} &	\textbf{Sent} \\
\hline
\hline
Attention BiRNN \cite{liu:2016}&	74.6 &	86.4 &	39.1 &	95.4 &	89.4 &	59.5\\
\hline
Slot-Gated \cite{Goo:2018}& 63.9&	87.7	&35.5	&94.6	&87.9	&55.4\\
\hline
Bi-Model \cite{wang:2018} & 70.3 & 	83.9 & 	34.4 & 	95.6 & 	90.7 & 	63.4\\
\hline
SF-ID \cite{Haihong:2019} & 66.2 & 	87.4 & 	34.9 & 	95.0 & 	90.6 & 	59.9\\
\hline
Stack-Propagation \cite{Qin:2019}& 72.1 & 	87.8 & 	40.1 & 	96.0 & 	94.2 & 	72.9\\
\hline
Joint Multiple ID-SF \cite{gangadharaiah-narayanaswamy:2019} &	73.4 &		84.6 &		36.1 &		95.1 &		90.6 &		62.9\\
\hline
AGIF \cite{qin:2020}& 74.4 & 	86.7 & 	40.8 & 	95.1 & 	94.2 & 	74.2\\
\hline
GL-GIN \cite{qin:2021}&76.3 &	88.3 &	43.5 &	95.6 &	94.9 &	75.4\\
\hline
DGM \cite{ding:2021}&76.7 &	88.7 &	47.1 &	96.7 &	94.7 &	78.0\\
\hline
SDJN \cite{chen:2022} &77.1 &	88.2 &	44.6 &	96.5 &	94.4 &	75.7\\
\hline
GISCo \cite{song:2022} &75.0 &	88.5 &	48.2 &	95.5 &	95.0&	75.9 \\
\hline
SSRAN \cite{cheng:2023} &77.9 &	89.4 &	48.9 &	\textbf{98.4} &	95.8&	77.5 \\
\hline
SLIM \cite{cai:2022}&78.3 &	88.5 &	47.6 &	97.2 &	96.5&	84.0 \\
\hline
TFMN \cite{chen-coling:2022}  & 79.8 &	88.0 &	50.2 &	97.7 &	96.4 &	84.7 \\
\hline
\hline
\textbf{Our model (BiSLU)} &	\textbf{81.5} &	\textbf{89.4} &	\textbf{51.5} & 97.8 & \textbf{97.2} & \textbf{85.4}\\\hline
\end{tabular}
\end{table*}

\begin{table*}[t]
\centering
\caption{Experimental results of different variants of our model. The best values for each column are shown in bold}
\begin{tabular}{|l|c|c|c|c|c|c|}
\hline
\multirow{2}{*}{\textbf{Model (Variant)}} & \multicolumn{3}{|c|}{\textbf{MixATIS}} &	\multicolumn{3}{|c|}{\textbf{MixSNIPS}} \\
\cline{2-7}
& \textbf{Intent} &	\textbf{Slot} & \textbf{Sent}  &	 \textbf{Intent} &	\textbf{Slot} &	\textbf{Sent} \\
	\hline
	\hline
 Using intermediate intents	&77.8 & 87.9 &	47.4 &	96.9	 & 96.4	 &	83.3\\
 \hline
Intermediate intent removal & 78.2  &	88.1  &	47.9 & 96.3	 &	96.5 & 84.5\\
\hline
Intent-slot attention & 79.1 &	88.5 &	48.7 &	96.9 & 96.2 & 84.7 \\
\hline
BiSLU Softmax &  78.9 &	87.5 &	47.3 &	95.8 &	95.5 & 83.7\\
\hline
BiSLU CRFs & 78.7 &	87.6 &	47.6 &	96.8 &	95.9 &	84.4 \\
\hline
\hline
\textbf{BiSLU Biaffine} & \textbf{79.5}	& 	\textbf{88.8}	& 	\textbf{49.2}	& 	\textbf{97.3}	& 	\textbf{96.5}	& 	\textbf{84.8} \\
\hline
\end{tabular}
\label{tab:Ablation}
\end{table*}

\begin{table*}[t]
\centering
\caption{Experimental results with different joint loss functions}
\label{tab:loss}
\begin{tabular}{|l|c|c|c|c|c|c|}
\hline
\multirow{2}{*}{\textbf{Model}} & \multicolumn{3}{|c|}{\textbf{MixATIS}} &	\multicolumn{3}{|c|}{\textbf{MixSNIPS}} \\
\cline{2-7}
& \textbf{Intent} &	\textbf{Slot} &	\textbf{Sent} & \textbf{Intent} &	\textbf{Slot} &	\textbf{Sent} \\
	\hline
	\hline
BiSLU (without SCL and self-distillation)	& 79.5	& 	88.8	& 	49.2	& 	97.3	& 	96.5	& 	84.8 \\
\hline
\hline
BiSLU  + intent SCL &	80.5 &	89.0 &	50.6 &	97.5 &	96.7 &	85.0 \\
\hline
BiSLU  + slot SCL &	80.4 &	89.2 &	50.3 &	97.4 &	 96.9 &	85.1 \\
\hline
BiSLU  + both SCL &	80.9 &	89.1 &	51.1 &	97.5 &	97.0 &	85.3 \\
\hline
\hline
\textbf{BiSLU + both SCL + self-distillation} &\textbf{81.5} &	\textbf{89.4} &	\textbf{51.5} & \textbf{97.8} & \textbf{97.2} & \textbf{85.4}\\
\hline
\end{tabular}
\end{table*}

\begin{figure*}        
	\begin{center}
		\includegraphics[width = 1\textwidth]{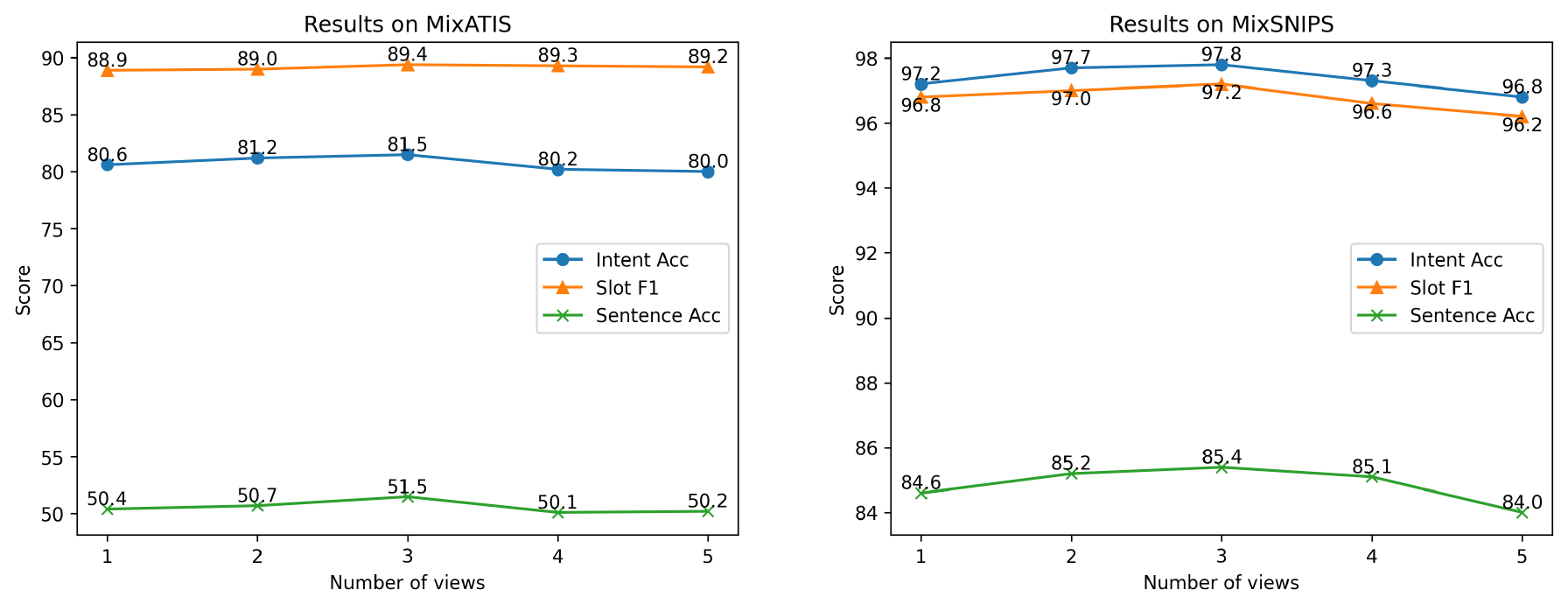}
	\end{center}
	\vspace{-15pt}
		\caption{Effects of the number of views.}
	\label{fig:n_views}
\end{figure*}

\subsection{Models to Compare}
We first conducted experiments to compare our proposed method with state-of-the-art multi-intent SLU models. For those models that were originally designed to address single-intent detection, including Attention BiRNN \cite{liu:2016}, Slot-Gated \cite{Goo:2018}, Bi-Model \cite{wang:2018}, SF-ID \cite{Haihong:2019} and Stack-Proagation \cite{Qin:2019}, we cited the results reported by Qin et al. \cite{qin:2020}. The comprehensive list of all the baseline models employed for the comparison is provided below:
\begin{itemize}
    \item Attention BiRNN \cite{liu:2016} : an alignment-based RNN model for joint slot filling and intent detection.
    \item Slot-Gated \cite{Goo:2018}: a unidirectional model that uses attention-based BiLSTMs with a slot-gated mechanism to leverage an intent context vector for improving slot filling.
    \item Bi-Model \cite{wang:2018}: a bidirectional model for single intent detection and slot filling.
    \item SF-ID \cite{Haihong:2019}: a bidirectional model with SF and ID subnets.
    \item Stack-Propagation \cite{Qin:2019}: a unidirectional model with stack-propagation which can directly use the intent information as an input for slot filling.
    \item Joint Multiple ID-SF \cite{gangadharaiah-narayanaswamy:2019}: a multi-task framework with a slot-gated mechanism for multiple intent detection and slot filling.
    \item AGIF \cite{qin:2020}: an adaptive interaction network to achieve fine-grained multi-intent information integration.
    \item GL-GIN \cite{qin:2021}: a non-autoregressive approach for joint multiple intent detection and slot filling.
    \item DGM \cite{ding:2021}: a dynamic graph model for joint multiple intent detection and slot filling.
    \item SDJN \cite{chen:2022}: a joint model for multi-intent SLU with self-distillation for slots.
    \item  GISCo \cite{song:2022}:  a graph neural network based on the global intent-slot co-occurrence to model the interaction between the two tasks.
    \item SSRAN \cite{cheng:2023}: a scope sensitive and result attentive model for multi-intent SLU based on Transformer.
    \item SLIM \cite{cai:2022}: a multi-intent SLU framework that uses a slot-intent classifier to learn the many-to-one mapping between slots and intents based on BERT. 
    \item TFMN \cite{chen-coling:2022}: a transformer-based threshold-free multi-intent SLU model.
\end{itemize}
\subsection{Experimental Setup}

In this work, we developed models utilizing the PyTorch framework in conjunction with the HuggingFace library\footnote{https://huggingface.co}. We employed the BERT base model\footnote{https://huggingface.co/bert-base-uncased} as pre-trained language models for our experimental purposes. Throughout all experiments, we established the maximum sequence length at $100$, while setting the dimensions of the output layers of the feed-forward start/end networks and biaffine networks, $k$ and $s$, to $300$ and $200$, respectively.

We trained the models using the AdamW optimizer \cite{Loshchilov:2019}, with default values for epsilon and weight decay in PyTorch (i.e., 1e-8). In order to ascertain the optimal hyper-parameters, we performed a grid search on the validation set, adjusting the AdamW initial learning rate within the range of \{1e-5, 2e5, 3e-5, 4e-5, 5e-5\}, the batch size within the range \{8, 16, 32\}, and the mixture weight $\lambda_i$ within the range \{0.05, 0.10, 0.15, \ldots, 0.80\}.  We tuned the intent thresshold $t_I$ and the number of views $V$ within the ranges \{0.3, 0.4, \ldots , 0.8\} and \{1,2,3,4,5\}, respectively.

For each model, we trained for $30$ epochs, subsequently evaluating intent accuracy, slot F$_1$ score, and sentence-level semantic frame accuracy on the validation set after each epoch. The model version that yielded the highest sentence-level semantic frame accuracy was ultimately selected to apply to the test set. 

\subsection{Experimental Results}
Table \ref{tab:Results} shows the performance of joint models for multiple intent detection and slot filling on the MixATIS and MixSNIPS datasets. The highest values in each column are highlighted in bold. It is evident from the table that our proposed model outperformed all the other models across all evaluation metrics on both datasets, except for the intent accuracy on the MixSNIPS dataset, where our model achieved the second-best result. Specifically, our model achieved 81.5\% intent accuracy, a slot F$_1$ score of 89.4\%, and 51.5\% sentence-level semantic frame accuracy on the MixATIS dataset. Compared with the second-best model, i.e., TMFN, our model improved 1.7\% (8.4\% error rate reduction), 1.4\% (11.7\% error rate reduction), and 1.3\% (2.6\% error rate reduction), respectively. On the MixSNIPS dataset, our model obtained 97.8\% intent accuracy, a slot F$_1$ score of 97.2\%, and 85.4\% sentence-level semantic frame accuracy, which improved 0.1\% (4.3\% error rate reduction), 0.8\% (22.2\% error rate reduction) and 0.7\% (4.6\% error rate reduction) compared with TMFN, respectively.

\subsection{Effects of Design Solutions}

Next, we conducted experiments with six variants of our model to evaluate the effects of the model’s design.
\begin{enumerate}
    \item \textbf{Using intermediate intents}. We used the intermediate intents as the final intents. 
    \item \textbf{Intermediate intent removal}. We removed the intermediate intents from the model architecture. 
    \item \textbf{Intent-slot attention}. Instead of directly using the intermediate intent probabilities, we fed them into an intent-slot attention layer as in \cite{Dao:2021}.
    \item \textbf{BiSLU CRFs}. We casted the slot filling task as a sequence labeling problem with conditional random fields (CRFs).
    \item \textbf{BiSLU Softmax}. We used the softmax function instead of using CRFs as in the previous variant. 
    \item \textbf{BiSLU Biaffine}. Our proposed model with the biaffine classifier. 
\end{enumerate}

Experimental results of the six variants on MixATIS and MixSNIPS datasets are shown in Table \ref{tab:Ablation}. Our proposed model BiSLU with the biaffine classifier achieved the best results on both datasets, demonstrating the reasonableness of our model architecture. The performance degradation of other variants also confirmed the importance of the intermediate intents, the slot classifier, and the final intents.  

To evaluate the efficacy of the proposed training method as well as the contribution of each type of loss, we investigated five variants of the training method:
\begin{enumerate}
    \item BiSLU without SCL and self-distillation: $\mathcal{L}=  \lambda_1 \mathcal{L}_{id} + \lambda_2 \mathcal{L}_{sf}$
    \item BiSLU with SCL for intents: $\mathcal{L}=  \lambda_1 \mathcal{L}_{id} + \lambda_2 \mathcal{L}_{sf} + \lambda_3 \mathcal{L}_{id\_scl}$
    \item BiSLU with SCL for slots:$\mathcal{L}=  \lambda_1 \mathcal{L}_{id} + \lambda_2 \mathcal{L}_{sf} +  \lambda_4 \mathcal{L}_{sf\_scl}$
    \item BiSLU with SCL for both intents and slots:$\mathcal{L}=  \lambda_1 \mathcal{L}_{id} + \lambda_2 \mathcal{L}_{sf} + \lambda_3 \mathcal{L}_{id\_scl} + \lambda_4 \mathcal{L}_{sf\_scl}$
    \item BiSLU with SCL for both intents and slots and self-distillation as well (our full method).   
\end{enumerate}

The results shown in Table \ref{tab:loss} demonstrated that incorporating both intent and slot contrastive losses significantly improved the model's performance. The BiSLU model with both contrastive losses achieved 80.9\% intent accuracy, 89.1\% in the slot F$_1$ score, and 51.1\% sentence-level semantic frame accuracy on the MixATIS dataset. Likewise, on the MixSNIPS dataset, the model got 97.5\% intent accuracy, 97.0\% in the slot F$_1$ score, and 85.3\% sentence-level semantic frame accuracy. The most significant performance enhancement was observed when using the joint loss function with all five types of losses: the intent loss, the slot loss, the intent contrastive loss, the slot contrastive loss, and the self-distillation loss. These findings suggested that incorporating supervised contrastive learning for both intent and slot and self-distillation led to a more robust and accurate model for multi-intent spoken language understanding.

\subsection{Effects of the Number of Views}
In contrastive learning, the procedure of generating positive and negative samples is essential and has a considerable effect on the performance of the learning models. To investigate the effect of the data augmentation method on the performance of our proposed model, we conducted experiments with a different number of views $V$. Figure \ref{fig:n_views} shows the experimental results of our joint models on the MixATIS and the MixSNIPS datasets, with the number of views varying from 1 to 5. The results demonstrated that our model was relatively stable and achieved the best results with $V = 3$ on the both datasets.    
\section{Conclusion}
We presented in this paper a bidirectional joint model for multiple intent detection and slot filling, two fundamental tasks in spoken language understanding. By predicting intermediate (soft) intents first, then slots, and the final intents, our model allows information to be transferred between the two tasks explicitly. We also introduced a novel training framework that includes supervised contrastive learning and self-distillation. Using a joint loss function consisting of different types of losses, i.e., intent loss, slot loss, intent contrastive loss, slot contrastive loss, and self-distillation loss between the intermediate and the final intents, the proposed model can be trained effectively. We empirically showed that our model outperformed state-of-the-art methods in both identifying intents and extracting slots on two benchmark datasets. Experimental results also demonstrated:  1) the reasonableness of our model’s design with the intermediate and the final intents as well as the using of a biaffine classifier for extracting slots; 2) the contribution of the contrastive learning and self-distillation components on the performance of the proposed model.   

\section*{Acknowledgement}
This work is supported by Posts and Telecommunications Institute of Technology, Hanoi, Vietnam.
\bibliography{ecai}
\end{document}